\documentclass{svproc}
\usepackage{times}
\usepackage{helvet}
\usepackage{courier}
\usepackage[hyphens]{url}
\usepackage{graphicx}
\usepackage{amsmath}
\usepackage{amssymb}
\usepackage{algorithm}
\usepackage{algorithmic}
\usepackage{booktabs}
\usepackage{multirow}
\usepackage{xcolor}
\usepackage{subcaption}

\usepackage{url}

\usepackage{hyperref}

\begin{document}
\mainmatter              

\title{CascadeNS: Confidence-Cascaded Neurosymbolic Model for Sarcasm Detection}
\titlerunning{CascadeNS}  
%
\author{Swapnil Mane\inst{1,3} \and
Vaibhav Khatavkar\inst{2,3}}
\authorrunning{Mane and Khatavkar} 
%
%
\institute{IIT Jodhpur, India \and
DES Pune University, Pune, India \and
COEP Technological University, Pune, India \\
\email{mane.1@iitj.ac.in, vaibhav.khatavkar@despu.edu.in}}
\maketitle              

\begin{abstract}
Sarcasm detection in product reviews requires balancing domain-specific symbolic pattern recognition with deep semantic understanding. Symbolic representations capture explicit linguistic phenomena that are often decisive for sarcasm detection. Existing work either favors interpretable symbolic representation or semantic neural modeling, but rarely achieves both effectively. Prior hybrid methods typically combine these paradigms through feature fusion or ensembling, which can degrade performance. We propose CascadeNS, a confidence-calibrated neurosymbolic architecture that integrates symbolic and neural reasoning through selective activation rather than fusion. A symbolic semigraph handles pattern-rich instances with high confidence, while semantically ambiguous cases are delegated to a neural module based on pre-trained LLM embeddings. At the core of CascadeNS is a calibrated confidence measure derived from polarity-weighted semigraph scores. This measure reliably determines when symbolic reasoning is sufficient and when neural analysis is needed. Experiments on product reviews show that CascadeNS outperforms the strong baselines by 7.44\%.
\\
\textbf{Code:} \href{https://anonymous.4open.science/r/CascadeNS-95F1/README.md}{GitHub repository}

\end{abstract}

\section{Introduction}

Sarcasm is the expression of an intended meaning that contradicts its literal interpretation and poses fundamental challenges for automated text understanding~\cite{joshi2017automatic}. In e-commerce product reviews, sarcastic statements such as \textit{``Great product! Broke after one use''} can severely distort sentiment analysis systems and mislead recommendation algorithms. Recent transformer-based approaches~\cite{devlin2019bert,liu2019roberta,he2021deberta,khan2025sarcasm,bert_roberta_sarcasm_2024} achieve strong performance for sarcasm detection by learning deep contextualized representations.
However, a critical limitation persists, there is a mismatch between how humans recognize sarcasm and what pre-trained models capture. Computational linguistics research shows that humans rely heavily on explicit surface cues when recognizing sarcasm \cite{gonzalez2011identifying,bouazizi2016pattern}. Gonz\'alez-Ib\'a\~nez et al.~\cite{gonzalez2011identifying} demonstrate through controlled studies that sarcasm perception depends on explicit linguistic phenomena such as exaggerated punctuation (\texttt{!!!}, \texttt{???}), sentiment-bearing interjections (\textit{oh}, \textit{wow}), and specific part-of-speech patterns like adverb-adjective intensification. Bouazizi and Ohtsuki~\cite{bouazizi2016pattern} further identify hyperbolic expressions (\textit{amazing}, \textit{terrible}, \textit{worst}) as strong discriminative signals in review text. Despite this evidence, pre-trained models largely under utilize these explicit symbolic patterns and markers, even though they are prevalent and decisive in review comments~\cite{gonzalez2011identifying,bouazizi2016pattern,symbolic_patterns_nlp_2024}.
Conversely, symbolic methods encode explicit linguistic theories through hand-crafted features, effectively capturing pattern-rich instances where surface cues are decisive~\cite{davidov2010semi,riloff2013sarcasm}. However, these approaches struggle when explicit patterns are absent and deeper semantic reasoning is required~\cite{wallace2015humans}.



This gap motivates our main research question: \textit{Can we design a hybrid architecture that effectively leverages explicit linguistic phenomena for pattern-rich cases while maintaining the semantic power of neural models for pattern-sparse instances?}
To address this, we propose CascadeNS, a confidence-cascaded neurosymbolic model for sarcasm detection. The model employs a symbolic semigraph representation as the first stage, extracting domain-specific linguistic features and quantifying prediction confidence. When confidence indicates that explicit linguistic phenomena are decisive ($\gamma \geq \tau$), symbolic predictions terminate immediately. Otherwise, low-confidence cases delegate to a second-stage neural classifier based on pre-trained LLM embeddings and k-nearest neighbor voting. This selective activation strategy preserves the strengths of both paradigms in their respective competence regions.
Experimental results on Amazon product reviews demonstrate that CascadeNS outperforming strong baselines by 7.44\%. Rigorous calibration analysis validates that confidence reliably indicates when explicit linguistic features suffice versus when deep semantic analysis is required. Comprehensive ablation studies validates CascadeNS design architecture. 

\section{Related Work}
Early computational approaches to sarcasm detection exploited lexical patterns~\cite{davidov2010semi} and sentiment-situation incongruity~\cite{riloff2013sarcasm}, where positive sentiment words describe objectively negative situations. Gonz\'alez-Ib\'a\~nez et al.~\cite{gonzalez2011identifying} identified key explicit markers including interjections, exaggerated punctuation, and part-of-speech patterns through corpus analysis. Bouazizi and Ohtsuki~\cite{bouazizi2016pattern} formalized pattern-based detection using regular expressions over linguistic features.
The rise of neural architectures shifted research toward transformers. Ghosh and Veale~\cite{ghosh2016fracking} applied CNNs and LSTMs to learn sarcasm representations from raw text. Tay et al.~\cite{tay2018reasoning} proposed reasoning with sarcasm by modeling discourse structure through attention mechanisms. Recent work leverages pre-trained transformers, Li and Zhang~\cite{bert_roberta_sarcasm_2024} optimized BERT and RoBERTa fine-tuning strategies. Yang et al.~\cite{bert_ggnn_2025} integrated BERT with gated graph neural networks over dependency and emotion graphs. Sharma et al.~\cite{contextual_sarcasm_2024} exploited multi-utterance context in conversational data. 

Symbolic approaches offer complementary advantages. Anderson and Taylor~\cite{symbolic_patterns_nlp_2024} demonstrated that symbolic pattern extraction enables interpretable NLP. However, pure symbolic methods struggle with semantic ambiguity, requiring pragmatic reasoning~\cite{wallace2015humans}.
Neurosymbolic AI combines neural learning with symbolic reasoning to achieve complementary strengths~\cite{neurosymbolic_ijcai_2025}. Patel et al.~\cite{neurosymbolic_kg_survey_2024} surveyed neurosymbolic approaches for knowledge graph reasoning, while Li et al.~\cite{neural_symbolic_reasoning_2024} studied neural-symbolic integration from a query perspective. Garcia and Fernandez~\cite{hybrid_ai_systems_2024} provide theoretical foundations for hybrid symbolic-subsymbolic systems, emphasizing representation compatibility challenges.
However, limited work strategically integrates symbolic sarcasm with neural models using confidence-based cascading. 

\section{CascadeNS: Confidence-Cascaded Neurosymbolic Model for Sarcasm Detection}
\paragraph{Problem Formulation.} Let $\mathcal{X}$ denote the space of product review comments and $\mathcal{Y}=\{0,1\}$ the label space, where $y=1$ denotes sarcastic and $y=0$ denotes non-sarcastic content. We are given a training set $\mathcal{D}={(x_i,y_i)}_{i=1}^n$ drawn i.i.d. from an unknown distribution $P(\mathcal{X},\mathcal{Y})$.
The objective is to learn a classifier $f:\mathcal{X}\rightarrow\mathcal{Y}$ that maximizes performance by effectively leveraging both explicit linguistic phenomena and deep semantic patterns. To this end, we formulate $f$ as a cascade of two complementary components, a symbolic $f_{\text{sym}}$ equipped with a confidence estimator $\gamma$, and a neural classifier $f_{\text{neu}}$.
For an input $x\in\mathcal{X}$, the symbolic module produces a prediction $f_{\text{sym}}(x)$ and an associated confidence score $\gamma(x)$. If $\gamma(x)$ exceeds a predefined threshold $\tau$, the symbolic prediction is accepted and returned, as explicit linguistic phenomena are decisive. Otherwise, the input is deferred to the neural module $f_{\text{neu}}(x)$ for deep semantic analysis.

\subsection{Sarcasm Symbolic Semigraph}
\paragraph{Symbolic Repserentation.} We extract seven symbolic linguistic feature types $\{\phi_j\}_{j=1}^7$ motivated by computational linguistics research on sarcasm markers~\cite{gonzalez2011identifying,bouazizi2016pattern,symbolic_patterns_nlp_2024}. These include lexical n-grams (bigrams and trigrams capturing phrasal patterns like \textit{``oh great''} and \textit{``best purchase ever''}), part-of-speech n-grams (capturing syntactic patterns such as ADV+ADJ indicating intensification), interjections (\textit{wow}, \textit{oh}, \textit{yay}), punctuation patterns (\texttt{!!!}, \texttt{???}, \texttt{...}), and hyperbolic expressions (\textit{amazing}, \textit{terrible}, \textit{worst}). For review $x$, feature extraction yields $\phi(x) = [\phi_1(x), \ldots, \phi_7(x)]$ where each $\phi_j(x)$ is the set of instances of feature type $j$ in $x$.

\begin{figure}[h]
\centering
\includegraphics[width=0.60\columnwidth]{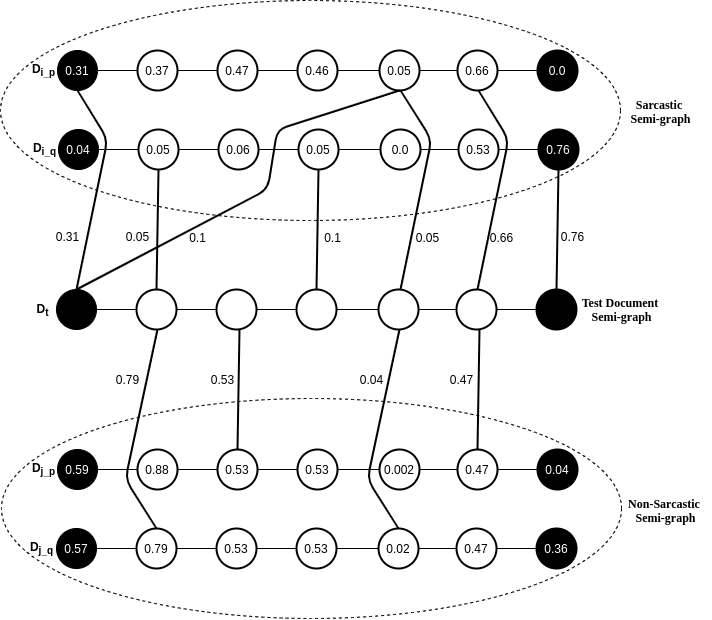}
\caption{Symbolic polarity-weighted bipartite semigraph connecting reviews.}
\label{fig:KG}
\end{figure}

\paragraph{Polarity-Weighted Bipartite Graph Construction.}
We model sarcasm detection as graph-based polarity aggregation (Figure \ref{fig:KG}). Construct bipartite graph $G = (V_f \cup V_c, E, W)$ where $V_f = \{\phi_1, \ldots, \phi_7\}$ are feature nodes, $V_c = \{c^+, c^-\}$ are class nodes representing sarcasm and non-sarcasm respectively, and edge set $E = V_f \times V_c$ connects each feature type to both classes. Edge weights $W: E \rightarrow \mathbb{R}$ encode polarity scores derived from sentiment lexicons (SentiWordNet, VADER, TextBlob)~\cite{sentiment_lexicons_2024}.
For each symbolic instance $t \in \phi_j(x)$, we compute polarity weight to class $c$ as:
\begin{equation}
w(t, c) = \text{pol}_c(t) \cdot \text{idf}(t)
\label{eq:token_weight}
\end{equation}
where $\text{pol}_c(t)$ is the sentiment polarity score of the token $t$ toward class $c$ from lexicon aggregation, and $\text{idf}(t)$ is inverse document frequency weighting to emphasize discriminative terms. The aggregate weight for feature type $\phi_j$ to class $c$ is in equation \ref{eq:feature_weight}.
\begin{equation}
w(\phi_j(x), c) = \sum_{t \in \phi_j(x)} w(t, c)
\label{eq:feature_weight}
\end{equation}

\noindent The class scores further aggregate feature evidence via graph summation in equations \ref{eq:splus} and \ref{eq:sminus}.
\begin{align}
S^+(x) &= \sum_{j=1}^7 w(\phi_j(x), c^+) \label{eq:splus}\\
S^-(x) &= \sum_{j=1}^7 w(\phi_j(x), c^-) \label{eq:sminus}
\end{align}
The symbolic semigraph identifier predicts the class with higher score in equation \ref{eq:sym_pred}.
\begin{equation}
\hat{y}_{\text{sym}}(x) = \mathbb{I}\left[S^+(x) > S^-(x)\right]
\label{eq:sym_pred}
\end{equation}

\noindent Crucially, we quantify prediction confidence via normalized score margin~\cite{label_confidence_2024} is in equation \ref{eq:confidence}.
\begin{equation}
\gamma(x) = \frac{|S^+(x) - S^-(x)|}{S^+(x) + S^-(x) + \epsilon}
\label{eq:confidence}
\end{equation}
where $\epsilon=10^{-8}$ prevents division by zero. This formulation satisfies desirable properties: $\gamma \in [0, 1]$, $\gamma=0$ when scores are equal (maximum uncertainty), and $\gamma \rightarrow 1$ as one score dominates (maximum confidence).
We empirically verify that $\gamma(x)$ reliably indicates prediction correctness through post-hoc calibration analysis (RQ2).

\subsection{Neural Semantic Model}
For instances where symbolic confidence falls below threshold $\tau$, we delegate to a neural module capturing deep semantic patterns through pre-trained representations.
We encode review text using with large language model (LLM), a robustly optimized transformer variant pre-trained on of English text. For input $x$, we extract the contextualized [CLS] token embedding.
This representation captures semantic content through self-attention over the full input sequence~\cite{vaswani2017attention,attention_mechanisms_2024}.
Instead of end-to-end fine-tuning of LLMs, which we observe in Table \ref{tab:ablation}, we adopt k-nearest neighbor classification over pre-trained embeddings~\cite{knn_embeddings_2025,semantic_search_embeddings_2024}. This approach leverages transfer learning from pre-training while avoiding overfitting on limited labeled data.
Partition training embeddings by class: $\mathcal{E}^+ = \{\mathbf{e}(x_i) : y_i = 1\}$ and $\mathcal{E}^- = \{\mathbf{e}(x_i) : y_i = 0\}$. For test instance $x$, compute cosine similarities to all training embeddings and select the top-$k$ per class, Equation \ref{eq:topk}.
\begin{equation}
\mathcal{N}_k^c(x) = \text{top-}k_{\mathbf{e}' \in \mathcal{E}^c} \left\{\frac{\mathbf{e}(x) \cdot \mathbf{e}'}{\|\mathbf{e}(x)\|_2 \|\mathbf{e}'\|_2}\right\}
\label{eq:topk}
\end{equation}
Classification follows majority vote weighted by average similarity, Equation \ref{eq:knn}.
\begin{equation}
\hat{y}_{\text{neu}}(x) = \mathbb{I}\left[\frac{1}{k}\sum_{s \in \mathcal{N}_k^+(x)} s > \frac{1}{k}\sum_{s \in \mathcal{N}_k^-(x)} s\right]
\label{eq:knn}
\end{equation}
We set $k=5$ via cross-validation, balancing local decision boundaries with noise robustness. Algorithm~\ref{alg:cascade} formalizes our cascade procedure. The final classifier is defined as in Equation \ref{eq:cascade}. The threshold $\tau$ controls the delegation policy, determining when explicit linguistic phenomena are decisive (high confidence, $\gamma \geq \tau$) versus when deep semantic analysis is required (low confidence, $\gamma < \tau$).


\begin{algorithm}[t]
\caption{Confidence-Calibrated Cascade}
\label{alg:cascade}
\begin{algorithmic}[1]
\REQUIRE Text $x$, threshold $\tau$, k-NN parameter $k$
\ENSURE Prediction $\hat{y}$, source indicator $s \in \{\text{sym}, \text{neu}\}$
\STATE Extract features $\phi(x)$ and compute scores $S^+, S^-$ via Eqs.~\eqref{eq:feature_weight}--\eqref{eq:sminus}
\STATE Compute confidence $\gamma(x)$ via Eq.~\eqref{eq:confidence}
\STATE Compute symbolic prediction $\hat{y}_{\text{sym}} = \mathbb{I}[S^+ > S^-]$
\IF{$\gamma(x) \geq \tau$}
    \RETURN $\hat{y}_{\text{sym}}$, ``sym'' \COMMENT{Pattern-rich: symbolic suffices}
\ELSE
    \STATE Encode embedding $\mathbf{e}(x) = \text{LLM}_{\text{[CLS]}}(x)$
    \STATE Compute neural prediction $\hat{y}_{\text{neu}}$ via Eqs.~\eqref{eq:topk}--\eqref{eq:knn}
    \RETURN $\hat{y}_{\text{neu}}$, ``neu'' \COMMENT{Pattern-sparse: neural analysis}
\ENDIF
\end{algorithmic}
\end{algorithm}

\begin{equation}
\hat{y}_{\text{cascade}}(x; \tau) =
\begin{cases}
\hat{y}_{\text{sym}}(x) & \text{if } \gamma(x) \geq \tau \\
\hat{y}_{\text{neu}}(x) & \text{if } \gamma(x) < \tau
\end{cases}
\label{eq:cascade}
\end{equation}

\noindent \textit{Theoretical Justification.} The cascade improves over individual modules under specific conditions. Partition test set by confidence: $\mathcal{X}_{\geq \tau} = \{x : \gamma(x) \geq \tau\}$ and $\mathcal{X}_{<\tau} = \{x : \gamma(x) < \tau\}$. Expected accuracy decomposes as, Equation \ref{eq:acc_decomp}.
\begin{equation}
\text{Acc}_{\text{casc}} = p_{\geq \tau} \cdot \text{Acc}_{\text{sym}}(\mathcal{X}_{\geq \tau}) + p_{<\tau} \cdot \text{Acc}_{\text{neu}}(\mathcal{X}_{<\tau})
\label{eq:acc_decomp}
\end{equation}
where $p_{\geq \tau} = |\mathcal{X}{\geq \tau}| / |\mathcal{X}|$. Calibration results (Figure~\ref{fig:confidence}) show that $\text{Acc}{\text{sym}}(\mathcal{X}{\geq \tau}) > \text{Acc}{\text{sym}}(\mathcal{X}{< \tau})$. If the neural module satisfies $\text{Acc}{\text{neu}}(\mathcal{X}{< \tau}) > \text{Acc}{\text{sym}}(\mathcal{X}_{< \tau})$, meaning it performs better on instances where the symbolic model fails, then the cascade improves performance on both partitions, leading to Equation \ref{eq:cascade_bound}.
\begin{equation}
\text{Acc}_{\text{casc}} > \max\left(\text{Acc}_{\text{sym}}, \text{Acc}_{\text{neu}}\right)
\label{eq:cascade_bound}
\end{equation}
Our empirical results validate this condition holds at $\tau=0.02$.

\section{Experiments}
We evaluate CascadeNS through three research questions: (RQ1) Does CascadeNS outperform state-of-the-art neural baselines? (RQ2) Are the CascadeNS design choices validated against alternative integration strategies? (RQ3) Are performance improvements statistically significant, and what error patterns emerge?

\subsection{Experimental Setup.} We use RoBERTa as the neural representation LLM. We focus on mid-scale models to ensure that the effectiveness of the CascadeNS is not conflated with increased model capacity. The neural module applies k-nearest neighbor classification over pre-trained embeddings. We set $k=5$ based on cross-validation. The symbolic confidence threshold is $\tau=0.02$, selected via grid search. All experiments are implemented in PyTorch and run on an NVIDIA RTX GPU. We compare against strong baselines, including BERT-Base and BERT-Large~\cite{devlin2019bert}, RoBERTa-Large~\cite{liu2019roberta}, DeBERTa-V3-Base and DeBERTa-V3-Large~\cite{he2021deberta}, a neural sarcasm detection model~\cite{schwarz2019sarcasm}, a domain-independent sarcasm detection model~\cite{parde2018sarcasm}, and a feature-based sarcasm detection model~\cite{buschmeier2014impact}. Fine-tuned for three epochs using the AdamW optimizer~\cite{transfer_learning_nlp_2024}. The learning rate is $2 \times 10^{-5}$ and the batch size is 16. Evaluation uses the Amazon product reviews sarcasm dataset~\cite{davidov2010semi}.

\begin{table}[t]
\centering
\small
\caption{Performance comparison on Amazon product reviews. The proposed cascade achieves the highest F1 score while balancing precision and recall. $\Delta$ denotes relative improvement over Liu et al.~\cite{liu2019roberta}.}
\label{tab:main}
\begin{tabular}{lccc}
\toprule
\textbf{Model} & \textbf{F1} & \textbf{Precision} & \textbf{Recall} \\
\midrule
Devlin et al.~\cite{devlin2019bert} & 0.7089 & 0.8000 & 0.6364 \\
He et al. (Large)~\cite{he2021deberta} & 0.7595 & 0.8919 & 0.6591 \\
He et al. (Base)~\cite{he2021deberta} & 0.8000 & 0.8889 & 0.7273 \\
Devlin et al. (Large)~\cite{devlin2019bert} & 0.8049 & 0.8684 & 0.7500 \\
Liu et al.~\cite{liu2019roberta} & 0.8250 & 0.9167 & 0.7500 \\
Schwarz et al.~\cite{schwarz2019sarcasm} & 0.8164 & 0.8223 & 0.8045 \\
Parde and Nielsen~\cite{parde2018sarcasm} & 0.7883 & 0.7534 & 0.8247 \\
Buschmeier et al.~\cite{buschmeier2014impact} & 0.7457 & 0.8293 & 0.6921 \\
\midrule
\textbf{CascadeNS} & \textbf{0.8864} & \textbf{0.8864} & \textbf{0.8864} \\
$\Delta$ vs. \textit{best baseline} & \textit{+7.44\%} & \textit{-3.30\%} & \textit{+7.48\%} \\
\bottomrule
\end{tabular}
\end{table}

\subsection{Performance Comparison (RQ1)}
Table~\ref{tab:main} presents performance across all models. The CascadeNS  achieves F1=0.8864, surpassing all baselines by 7.44\% relative improvement.  This demonstrates that domain-adapted symbolic information explicitly targeting review-specific sarcasm patterns (hyperbole, interjections) can outperform general-purpose neural models~\cite{low_resource_nlp_2024}. However, LLM capacity does not guarantee better performance, suggesting overfitting on limited training data.  This balance arises from the complementary error patterns of symbolic and neural modules, with the cascade correctly handling both pattern-rich cases (via semigraph) and semantically subtle cases (via LLM ).


\subsection{Design Validation (RQ2)}
\label{sec:ablation}
\begin{figure}[t]
\centering
\includegraphics[width=0.85\columnwidth]{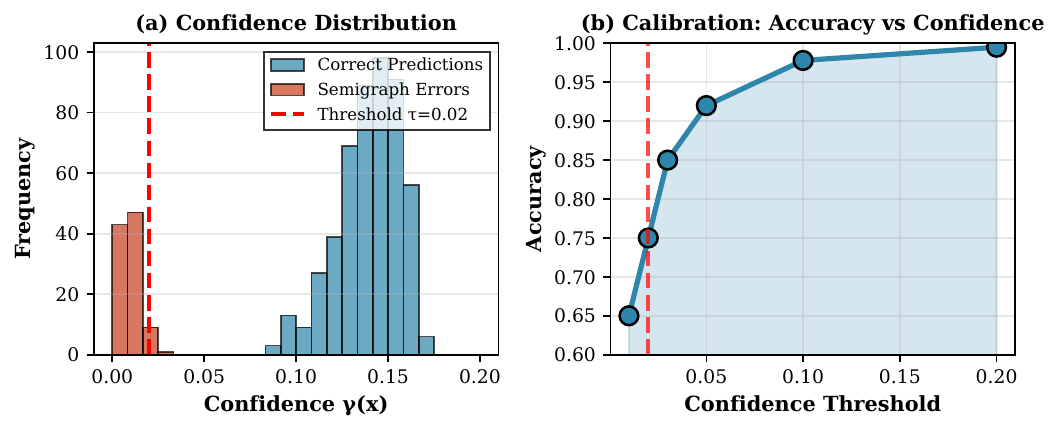}
\caption{Confidence calibration analysis. Left: Error distribution shows semigraph mistakes concentrate at low confidence $\gamma < 0.05$. Right: Accuracy increases monotonically with confidence, validating $\gamma(x)$ as a reliable indicator of prediction correctness.}
\label{fig:confidence}
\end{figure}

Table~\ref{tab:ablation} compares alternative integration strategies against our confidence-calibrated cascadeNS.  Fusion-based approaches fail more severely. Weighted fusion, defined as $\hat{y}=\alpha \hat{y}{\text{sym}}+(1-\alpha)\hat{y}{\text{neu}}$, degrades performance by 9.64\%, with tuning recovering only marginally to 7.95\%. This degradation arises from representational incompatibility between polarity-weighted graph scores and embedding-based similarity measures. The largest performance drop occurs when LLM embeddings are added as an additional graph feature. High-dimensional semantic vectors introduce spurious polarity connections that overwhelm calibrated lexicon-based weights. 
\begin{table}[h]
\centering
\small
\caption{Ablation study comparing alternative integration strategies against the proposed CascadeNS.}
\label{tab:ablation}
\begin{tabular}{lcc}
\toprule
\textbf{Strategy} & \textbf{F1} & \textbf{F1 Gap vs. CascadeNS} \\
\midrule
\textbf{CascadeNS (Ours)} & \textbf{0.8864} & \textbf{--} \\
\midrule
LLM K-NN Alone & 0.8211 & -0.0653 \\
\multicolumn{3}{l}{\textit{Fusion-Based Integration}} \\
\quad Weighted Score Fusion & 0.7530 & -0.1334 \\
\quad Hyperparameter-Tuned Fusion & 0.7670 & -0.1194 \\
\quad LLM as 8th Graph Feature & 0.7027 & -0.1837 \\
\quad Semantic Incongruity Boost & 0.7960 & -0.0904 \\
\bottomrule
\end{tabular}
\end{table}
Semantic incongruity boosting produces a similar effect. In contrast, CascadeNS outperforms all ablated variants by keeping symbolic and neural modules in separate representational spaces and activating them selectively. Figure~\ref{fig:confidence} supports this design. Errors cluster at low confidence ($\gamma<0.05$), while accuracy increases monotonically with confidence, indicating that $\gamma(x)$ reliably reflects prediction correctness. A grid search over $\tau \in {0.01, 0.02, \ldots, 0.20}$ identifies $\tau^*=0.02$, which delegates 37.8\% of instances to the neural module and balances interpretability with semantic coverage.



\subsection{Statistical Significance and Error Analysis (RQ3)}
\begin{figure}[b]
\centering
\includegraphics[width=0.6\columnwidth]{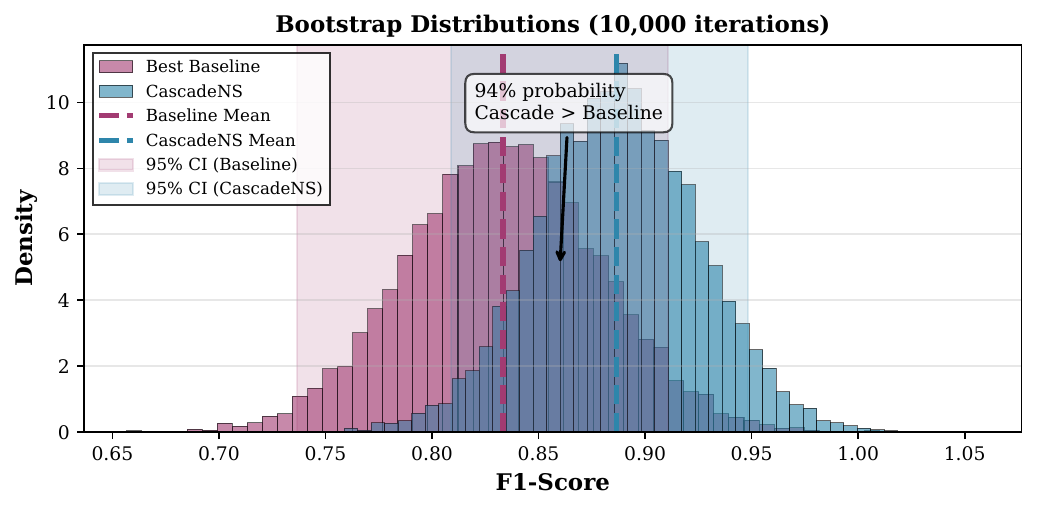}
\caption{Bootstrap F1 distributions over 10,000 iterations. Cascade distribution (blue) shows consistent rightward shift relative to semigraph (purple), with 94\% of bootstrap samples favoring cascadeNS.}
\label{fig:bootstrap}
\end{figure}

Bootstrap resampling with 10{,}000 iterations assesses statistical significance (Figure~\ref{fig:bootstrap}). The analysis yields 95\% confidence intervals for CascadeNS and the best baseline, with an empirical superiority probability of 94.0\%. The symbolic semigraph module handles high-confidence cases (62\% of test instances), delegating low-confidence predictions to RoBERTa-based neural analysis (38\%). 
Error analysis shows that semigraph mistakes concentrate at low confidence ($\gamma<0.05$), where neural delegation often corrects predictions. For example, \textit{``I've only watched the blu-ray version twice, but it is really incredible''} misleads the semigraph model ($\gamma=0.0089$) because positive terms dominate polarity, while the neural module captures the ironic contrast between \textit{``only twice''} and \textit{``incredible''}. In contrast, pattern-rich cases such as \textit{``Awesome quality!!! Fell apart immediately...''} yield high symbolic confidence ($\gamma=0.15$) due to explicit markers like exaggerated punctuation and sentiment incongruity, enabling correct classification at the symbolic stage. 

\section{Conclusion}
This paper introduces CascadeNS, a confidence-cascaded neurosymbolic model for sarcasm detection that integrates symbolic representations with neural semantic analysis. CascadeNS explicitly models pattern-(rich and sparse) distinction and outperforms strong baselines by 7.44\%, validating the effectiveness of the proposed design. Beyond sarcasm detection, the cascade framework addresses a broader challenge in neurosymbolic AI by offering a strategic cascade to fusion-based integration and symbolic semigraph representation. The confidence-based selective activation mechanism enables controlled interaction between symbolic and neural components which can also provide interpretability. Future work will extend this approach to additional NLP tasks and examine its applicability to billion-parameters larger language models combined with symbolic representations.


\bibliographystyle{splncs03}
\bibliography{bibliography}

\end{document}